# Probabilistic Belief Change:
# Expansion, Conditioning and Constraining

**Frans Voorbraak**
Dept. of Mathematics and Computer Science
University of Amsterdam
Plantage Muidergracht 24
1018 TV Amsterdam, The Netherlands

## Abstract

The AGM theory of belief revision has become an important paradigm for investigating rational belief changes. Unfortunately, researchers working in this paradigm have restricted much of their attention to rather simple representations of belief states, namely logically closed sets of propositional sentences. In our opinion, this has resulted in a too abstract categorisation of belief change operations: expansion, revision, or contraction. Occasionally, in the AGM paradigm, also probabilistic belief changes have been considered, and it is widely accepted that the probabilistic version of expansion is conditioning. However, we argue that it may be more correct to view conditioning and expansion as two essentially different kinds of belief change, and that what we call constraining is a better candidate for being considered probabilistic expansion.

## 1 Introduction

The AGM theory of belief revision (Gärdenfors 1988, 1992) has become an important paradigm for investigating rational belief changes. In this theory, three main types of belief changes are distinguished, namely expansion, revision, and contraction. Expansion is the most simple type of belief change, partly because it is supposed to occur only when information is added which is consistent with the previously held beliefs, whereas the other types of belief changes (also) apply in case the new information is inconsistent with the old beliefs.

In fact, if logical theories are used to represent the belief states, then an explicit definition of expansion can be given: The result of expanding a theory $K$ with a sentence $\phi$ is the set $Cn(K \cup \{\phi\})$ of logical consequences of $K \cup \{\phi\}$. In general, no such explicit definition can be given for revision and contraction.

When these types of belief changes are studied in the context of probabilistic belief states, at first sight, conditioning seems to be the obvious probabilistic variant of expansion. We argue that closer inspection shows that conditioning and expansion are best viewed as two essentially different types of belief changes. This is most clear in the context of partial probability theory, where both types of belief changes can be compared.

In our view, conditioning is a type of belief change different from expansion, revision, and contraction. Conditioning does not make sense in the context of belief states represented by logical theories, just as expansion does not make sense in the context of belief states represented by probability functions. Expansion and conditioning both make sense in the context of belief states represented by partially determined probability functions, since they allow the representation of both ignorance and uncertainty.

In the remainder of this paper, we first review several models of belief states, including belief sets and (partial) probabilistic models. Next, we discuss the notions of expansion and conditioning in the different contexts, and we provide several arguments for our opinion that these notions are essentially different, and that constraining is better suited than conditioning to be considered probabilistic expansion. In (Voorbraak, 1996), we briefly discuss the probabilistic variant of revision, but in this paper we restrict ourselves to expansion. We conclude with a discussion of the question how to determine whether either conditioning or constraining is appropriate.

## 2 Belief State Models

Throughout the paper, $L$ denotes a propositional language with the usual connectives $\neg$, $\vee$, $\wedge$, $\rightarrow$, $\leftrightarrow$, and constants $\bot$ and $\top$. For simplicity, we assume that $L$ has finitely many propositional letters $p_1, p_2, \ldots, p_n$. We write $SL$ for the set of sentences of $L$ and we use $\phi$, $\psi$, ... as sentential variables.

We further assume $\vdash$ to denote the standard propositional consequence relation on $SL$ and Cn to denote the associated consequence operation. (If $S \subseteq SL$, then $Cn(S) = \{\phi \in SL : S \vdash \phi\}$.) A set $S \subseteq SL$ is called *logically closed* iff $S = Cn(S)$.

**Definition 1 (belief set)** *A belief set $K$ for $L$ is a logically closed consistent subset of $SL$. A sentence $\phi$ of $L$ is called accepted in $K$ iff $\phi \in K$.*

For technical convenience, the inconsistent belief state (containing all sentences of the language) is sometimes added to the considerations. However, in this paper, the rational belief states are assumed to be consistent.

A sentence $\alpha \in SL$ is called an *atom* of $L$ if $\alpha = \pi_1 \wedge \pi_2 \wedge \ldots \wedge \pi_n$, where for each $i \in \{1, \ldots, n\}$, $\pi_i = p_i$ or $\pi_i = \neg p_i$. Notice that if $n$ is the number of propositional letters of $L$, then the number $N$ of atoms of $L$ equals $2^n$. A belief set $K$ for $L$ can be represented equivalently by the set of all atoms of $L$ which are consistent with $K$. Since atoms can be identified with possible worlds, this is called a possible worlds representation of belief states.

The belief sets and the possible worlds models are essentially equivalent, but there exist many other essentially different models of belief states. For example, probabilistic models can be used to represent *degrees* of belief in propositions, as opposed to belief sets which only model whether a proposition is (fully) accepted or not.

**Definition 2 (probabilistic model)** *A probabilistic (belief) model $P$ for $L$ is a probability function on $SL$, that is, $P$ is a function on $SL$ satisfying the following three conditions.*

1. *For all $\phi \in SL$, $P(\phi) \geq 0$*

2. *For all $\phi \in SL$, if $\vdash \phi$, then $P(\phi) = 1$*

3. *For all $\phi, \psi \in SL$, if $\vdash \neg(\phi \wedge \psi)$, then $P(\phi \vee \psi) = P(\phi) + P(\psi)$.*

*A sentence $\phi$ of $L$ is called accepted in $P$ iff $P(\phi) = 1$.*

We write $PROB(SL)$ for the class of probability functions on $SL$. With every probabilistic model $P$ for $L$ one can naturally associate the belief set $t(P)$ for $L$ given by $t(P) = \{\phi \in SL : P(\phi) = 1\}$. It is easy to see that $t(P)$ is indeed a belief set and that in $t(P)$ the same sentences are accepted as in $P$. The belief set $t(P)$ is called the *top* of $P$. Of course, different probability functions may have the same top. The belief sets for $L$ correspond exactly to the equivalence classes of probabilistic models for $L$ with the same top.

In other words, with every belief set $K$ for $L$ one can naturally associate a class of probability functions. Since also a probabilistic belief model $P$ for $L$ can be naturally associated with a class of probability functions, namely $\{P\}$, it follows that using classes of probability functions as models of belief states naturally incorporates the previously defined models.

**Definition 3 (partial probabilistic model)**
*A partial probabilistic model $\Pi$ for $L$ is a non-empty class of probability functions on $SL$. A sentence $\phi$ of $L$ is called accepted in $\Pi$ iff for every $P \in \Pi$, $P(\phi) = 1$.*

Let $i$ be the function embedding belief sets and probability models into partial probability models in the way mentioned above. That is, $i$ assigns to a belief set $K$ for $L$, or probability function $P$ on $SL$, the class of probability functions on $SL$ compatible with $K$, or $P$. Thus, $i(K) = \{P \in PROB(SL) : \text{if } \phi \in K, \text{then } P(\phi) = 1\}$, and $i(P) = \{P\}$.

The previously introduced function $t$ for taking the top of a probability function can naturally be extended to partial probabilistic models as follows. If $\Pi$ is a class of probability functions on $SL$, then $t(\Pi) = \{\phi \in SL : \text{for every } P \in \Pi, P(\phi) = 1\}$. Notice that $t(i(P))$ agrees with $t(P)$ as previously defined, and that $t(i(K)) = K$.

Partial probabilistic models are of technical interest, since they generalise both the belief sets and the probabilistic models. In addition, it can be argued that in some situations, in particular when there is ignorance concerning the exact likelihood of events, a class of probability functions is more appropriate to model a belief state than a single probability function. Classes of probability functions are also mentioned in (Gärdenfors, 1988) as possible belief state models.

We conclude this section by pointing out that a partial probabilistic model can be viewed as some kind of possible worlds model for a probabilistic language. Let $L_{PROB}$ be a language for probabilistic reasoning with $L$ as object language for the probability expressions. We use $SL_{PROB}$ to refer to the sentences of $L_{PROB}$, and $\vdash_{PROB}$ and $Cn_{PROB}$ denote a probabilistic consequence relation and associated consequence operation on $SL_{PROB}$. We assume that the sentences of $L$ are not sentences of $L_{PROB}$, but occur only in the scope of the probability operator of $L_{PROB}$. For example, one can take $L_{PROB}$ to be (a suitable adaptation of) $L(AX)$ or $L(AX_{FO})$ of (Fagin et al., 1988), and use $AX_{MEAS}$ or $AX_{FO-MEAS}$ of the same paper as the probabilistic logic.

Sentences of $L_{PROB}$ can be viewed as constraints on probability functions on $SL$, much as sentences of $L$ can be viewed as constraints on possible worlds for $L$. Thus, any set $S \subseteq SL_{PROB}$ determines a partial probabilistic model for $L$. Of course, $S$ and $Cn_{PROB}(S)$ determine the same model. Hence there is an exact correspondence between (probabilistic) belief sets for $L_{PROB}$ and classes of probability functions determined by subsets of $SL_{PROB}$. For any such class $\Pi$, let $\langle \Pi \rangle$ denote the belief set $S$ for $L_{PROB}$ such that $\Pi = \{P \in PROB(SL) : P \text{ satisfies } S\}$. We will sometimes use $P \models S$ as an abbreviation for '$P$ satisfies $S$'.

Although, in general, a class of probability functions cannot be expected to correspond to a (probabilistic) belief set for $L_{PROB}$, it seems likely that any natural occurring class is determined by a set of sentences of some probabilistic language. Therefore, a probabilistic belief set may be an interesting representation of a belief state.

## 3    Conditioning and Constraining

Let $K$ be a belief set for $L$ and let $\phi \in SL$ be consistent with $K$, or, in other words, $\neg \phi$ not accepted in $K$. Then the belief in $\phi$ can be added to the belief state $K$ without giving up any old beliefs. In the AGM theory, such a kind of change in belief state is called an expansion, and is modelled by the operation + given by $K + \phi = Cn(K \cup \{\phi\})$.

If we use a probability function $P$ on $SL$ as our representation of belief state, then conditioning seems a natural candidate for the role played by the operation + on belief sets.

**Definition 4 (conditioning)** *Let $P$ be a probability function on $SL$. Define, for each $\phi \in SL$ such that $P(\phi) \neq 0$, the function $P_\phi$ as follows.*

> *For every $\psi \in SL, P_\phi(\psi) = P(\psi|\phi) = \dfrac{P(\phi \wedge \psi)}{P(\phi)}$.*

It is easy to check that the thus defined function $P_\phi$ is a probability function on $SL$. The mapping from $P$ to $P_\phi$ is strongly related to expansion with $\phi$, since, just as the expansion operation, the mapping is defined whenever $\neg \phi$ is not accepted in the original belief state ($P(\neg \phi) \neq 1$), and we have $t(P) + \phi = t(P_\phi)$. It follows that the expansion operation + on belief sets can be viewed as an abstraction of Bayesian conditioning on probabilistic belief states.

However, the expansion operation on belief sets can be viewed as an abstraction of many other operations on probabilistic belief models as well, including preservative imaging introduced in (Gärdenfors, 1988). The question arises whether compatibility with + is sufficient for an operation on probabilistic belief models to qualify as a probabilistic expansion. This question is perhaps even more relevant in the context of partial probabilistic belief models, where even more operations are compatible with the expansion of belief sets. The following definitions describe two such operations.

**Definition 5 (extended conditioning)** *Let $\Pi$ be a partial probabilistic model for $L$, and let $\phi \in SL$, such that for some $P \in \Pi$, $P(\phi) > 0$. Define the partial probabilistic model $\Pi_\phi$ for $L$ as follows.*

$$\Pi_\phi = \{P_\phi : P \in \Pi, P(\phi) > 0\}.$$

**Definition 6 (constraining)** *Assume that $\Pi$ is a partial probabilistic model for $L$ and that $\phi \in SL$, such that for some $P \in \Pi$, $P(\phi) = 1$. Define the partial probabilistic model $\Pi_{\&\phi}$ for $L$ as follows.*

$$\Pi_{\&\phi} = \{P \in \Pi : P(\phi) = 1\}.$$

These, or similar operations have been studied before. See, for example, (Dubois and Prade, 1997). In (Grove and Halpern, 1998), both conditioning and constraining, and several other ways of updating sets of probability measures, are discussed from an axiomatic point of view.

The following proposition shows that constraining and extended conditioning are both compatible with + on embeddings of belief sets.

**Proposition 1** *Let $\Pi$ be a partial probabilistic belief model for $L$ such that $\Pi = i(K)$, for some belief set $K$ for $L$, and let $\phi \in SL$ such that $\neg \phi \notin K$. Then $t(\Pi_{\&\phi}) = t(\Pi_\phi) = t(\Pi) + \phi$.*

Although both operations are compatible with expansions of belief sets, we consider constraining to be the proper probabilistic notion of expansion, whereas (extended) conditioning is in our opinion best viewed as a new kind of operation, different from both expansion. (It is of course also different from the other AGM operations, revision and contraction, since these are intended for incorporating new information which is inconsistent with the previously held beliefs.) In the following section, we give several arguments supporting this view.

## 4    Conditioning versus Expansion

We argue that constraining and not conditioning should be considered to be probabilistic expansion, since

1.  constraining, and not conditioning, can be viewed as expansion of probabilistic belief sets

2. constraining can be said to (primarily) reduce ignorance, just like expansion, whereas conditioning is (primarily) connected with reducing uncertainty.

This latter difference between conditioning and constraining will only be discussed briefly in general terms, but a concrete manifestation of this difference will be treated in more detail: like expansion, and in contrast to conditioning, constraining can be used to obtain any belief state from a state of ignorance while preserving intermediate results.

### 4.1   Expansion of Probabilistic Belief Sets

We argue, that, in contrast to conditioning, constraining can be viewed as expansion of probabilistic belief sets. First notice that constraining can also be defined for sentences of $L_{PROB}$ instead of $L$.

**Definition 7 (generalised constraining)** *Let $\Pi$ be a non-empty class of probability functions on SL. Define, for each $\phi \in SL_{PROB}$ which is satisfied by some $P \in \Pi$,*

$$\Pi_{\&\phi} = \{P \in \Pi : P \models \phi\}.$$

This definition of constraining generalises definition 6, since $\Pi_{\&\phi}$, for $\phi \in SL$, can be viewed as an abbreviation of $\Pi_{\&(P(\phi)=1)}$. Further on, we will also discuss an analogous generalisation of conditioning. The following proposition shows that the generalised version of constraining translates into the expansion of probabilistic belief sets.

**Proposition 2** *Assume that $S$ is a probabilistic belief set for $L_{PROB}$. Let $\Pi$ be the partial probabilistic belief model for $L$ such that $\langle\Pi\rangle = S$, and let $\phi \in SL_{PROB}$ be consistent with $S$. Then $\langle\Pi_{\&\phi}\rangle = Cn_{PROB}(S \cup \{\phi\})$.*

**Corollary 3** *Let $\Pi$ be a partial probabilistic belief model for $L$ such that $\langle\Pi\rangle = S$, and let $\phi \in SL$ such that for some $P \in \Pi$, $P(\phi) = 1$. Then $\langle\Pi_{\&\phi}\rangle = Cn_{PROB}(S \cup \{P(\phi) = 1\})$.*

The following example shows that the analogue of this corollary for conditioning does not hold, provided $Cn_{PROB}$ is monotone, and $L_{PROB}$ is nontrivial, in the sense that it allows assigning (probability) values between 0 and 1.

**Example 1** *Let $p$ and $q$ be the proposition letters of $L$, and let $\Pi$ be the partial probabilistic belief model for $L$ such that $\langle\Pi\rangle = Cn_{PROB}(\{P(p \wedge q) = x, P(p \wedge \neg q) = 0\})$, for some $x$ between 0 and 1. Then $\langle\Pi\rangle$ contains $P(p) = x$, whereas $\langle\Pi_q\rangle$ does not.*

Conditioning is not just a matter of adding information and possibly sharpening the bounds on probability values. Conditioning may also involve the revision of previously held degrees of belief, even if one conditions on events which are completely consistent with the old belief state. The possibility that conditioning is perhaps not a 'pure' expansion process since it has some aspects of a revision process has already been mentioned in (Dubois and Prade, 1992).

### 4.2   Reducing Uncertainty or Ignorance

We claim that constraining can be said to (primarily) reduce ignorance, whereas conditioning is (primarily) connected with reducing uncertainty. To make this more precise, one needs measures of both uncertainty and ignorance, preferably in the contexts of partially specified probability. Obtaining and justifying such measures is not easy, although some work has been done in this area. See, for example, (Klir, 1994) for a discussion of such measures in the context of Dempster-Shafer theory.

In (Voorbraak, 1996) we propose provisional measures for uncertainty and ignorance in the partial probabilistic case. The uncertainty measure is based on entropy and the ignorance measure is based on the (average) difference between upper and lower probability of events.

These measures are provisional, and cannot be justified rigorously, but they suffice to show that conditioning is biased towards reducing uncertainty, whereas constraining is biased towards reducing ignorance. More precisely, if the uncertainty in a belief state is not minimal, then it can always be reduced by conditioning, but not always by constraining. If the ignorance in a belief state is not minimal, then it can always be reduced by both constraining and conditioning, but constraining always (weakly) reduces the ignorance, whereas after conditioning the ignorance might be increased. See (Voorbraak, 1996) for details.

Since the expansion operation + on belief sets is aimed at reducing ignorance rather than uncertainty, the above can be viewed as a second argument in favour of the position that constraining rather than conditioning is the probabilistic variant of expansion. Below, we will consider a more specific version of this argument.

### 4.3   Expanding from Ignorance

Any belief set $K$ for $L$ can be obtained from the ignorant belief state $Cn(\emptyset)$ by a sequence of expansions. (In fact, since $L$ has a finite number of proposition letters, a single expansion suffices.) This is in accordance with the intuition that one can learn about a subject of which one is completely ignorant, without having to give up previously held beliefs.

Intuitively, one should be able to learn about a subject in bits and pieces, possibly getting information from different sources and on different occasions. Indeed, starting from the ignorant belief state $Cn(\emptyset)$, iterated expansion leads to more and more extended belief states. This is a monotone process, since the following preservation property holds. Let $K$ be a belief set for $L$, and let $\psi, \phi \in SL$ such that $K, \phi$, and $\psi$ are jointly consistent, then

$$\phi \text{ is accepted in } (K + \phi) + \psi. \quad (1)$$

Since for every probability function $P$ on $SL$ and $\phi, \psi \in SL$, such that $P(\phi \wedge \psi) > 0$, we have $(P_\phi)_\psi(\phi) = 1$, the analogous property holds in the context of probabilistic belief models. Here the most obvious candidate to represent the ignorant belief state is the uniform probability function $P_{un}$, defined by $P_{un}(\alpha) = \frac{1}{N}$, where $N$ is the number of atoms of the language, and $\alpha$ is any one of these atoms. The following example shows that it is *not* the case that any probability function $P$ on $SL$ can be obtained by conditioning the uniform probability function $P_{un}$ on $SL$.

**Example 2** *Let $p$ be the only proposition letter of $L$, and let $P$ be the probability function on $SL$ given by $P(p) = 0.1$. Then $P$ differs from $(P_{un})_p$, $(P_{un})_{\neg p}$, and $(P_{un})_{p \vee \neg p} (= P_{un})$, which are the only probability functions that can be obtained by conditioning $P_{un}$.*

This negative result can be circumvented if one generalises the notion of (Bayesian) conditioning by allowing conditioning on events which are not certain. Jeffrey conditioning is such a generalisation of Bayesian conditioning. Below we first define a simple version of Jeffrey conditioning, which we call binary Jeffrey conditioning, for reasons that will become clear further on.

**Definition 8 (binary Jeffrey conditioning)** *Let $P$ be a probability function on $SL$, and let $\phi \in SL$ such that $0 < P(\phi) < 1$. Define, for any $x \in [0,1]$, the function $P_{\phi,x}$ as follows.*

$$P_{\phi,x} = xP_\phi + (1-x)P_{\neg\phi}$$

Notice that the usual conditional probability function $P_\phi = P_{\phi,1}$. It is easy to see that the probability function of $P$ example 2 can be obtained from $P_{un}$ by binary Jeffrey conditioning: $P = (P_{un})_{p,0.1}$. Binary Jeffrey conditioning can be generalised in a natural way as follows.

**Definition 9 (Jeffrey conditioning)** *Let $P$ be a probability function on $SL$, and let $\{\phi_i : i \in I\}$ be*

a set of mutually exclusive sentences of $L$, such that for every $i \in I$, $P(\phi_i) > 0$. Define, for any set $\{x_i : i \in I\}$, with $x_i \in [0,1]$ and $\sum_{i \in I} x_i = 1$, the function $P_{\{(\phi_i, x_i):i \in I\}}$ as follows.

$$P_{\{(\phi_i, x_i):i \in I\}} = \sum_{i \in I} x_i P_{\phi_i}.$$

Notice that $P_{\phi,x} = P_{\{(\phi,x),(\neg\phi,1-x)\}}$. Hence binary Jeffrey conditioning is Jeffrey conditioning on *two* exclusive events. Bayesian conditioning is Jeffrey conditioning on a single event and might be called unary Jeffrey conditioning. Jeffrey conditioning can be generalised even further to allow conditioning on constraints expressed by sentences of $L_{PROB}$ as follows.

**Definition 10 (minimum cross entropy)** *Assume that $\alpha_1, \alpha_2, \ldots, \alpha_N$ are the atoms of $L$. Let $P$ be a probability function on $SL$, and let $\phi \in SL_{PROB}$. We define $P_\phi$, the minimum cross entropy update of $P$ with $\phi$, to be that probability function $P'$ on $SL$ satisfying $\phi$ where the function*

$$I(P', P) = \sum_{i=1}^{N} P'(\alpha_i) \log \frac{P'(\alpha_i)}{P(\alpha_i)}$$

*is minimal.*

The function $(P_{un})_\phi$ is the probability function satisfying $\phi$ with the maximum entropy. It follows that if $\phi$ uniquely determines a probability function $P$ ($P' \models \phi$ iff $P' = P$), then $(P_{un})_\phi = P$.

It is also easy to show that every probability function $P$ on $SL$ can already be obtained from $P_{un}$ by Jeffrey conditioning. In fact, the following proposition shows that one does not have to start from the 'ignorant' $P_{un}$, since (even binary) Jeffrey conditioning allows many probabilistic belief states to be changed into an arbitrary probabilistic belief state.

**Proposition 4** *Let $P$ be a probability function on $SL$ and let $\{\alpha_i : i \in I\}$ be the set of atoms of $L$ such that $P(\alpha_i) > 0$. Assume that $P'$ is a probability function on $SL$ such that for every $i \in I$, $P'(\alpha_i) > 0$. Then $P$ can be obtained from $P'$ by at most $|I|$ applications of binary Jeffrey conditioning.*

**Example 3** *Let $p$ and $q$ be the proposition letters of $L$, and let $P$ be the probability function on $SL$ given by the table below. Let $P' = P_{un}$. Then $P^1 = (P_{un})_{p \vee q}$, $P^2 = P^1_{p \wedge q, \frac{3}{4}}$, and $P^3 = P^2_{p, 0.9} = P$. The different probability functions are described in the table below.*

|        | $p \wedge q$ | $p \wedge \neg q$ | $\neg p \wedge q$ | $\neg p \wedge \neg q$ |
|--------|------|------|------|------|
| $P$    | 0.4  | 0.5  | 0.1  | 0    |
| $P' = P_{un}$ | 0.25 | 0.25 | 0.25 | 0.25 |
| $P^1$  | $\frac{1}{3}$ | $\frac{1}{3}$ | $\frac{1}{3}$ | 0 |
| $P^2$  | $\frac{2}{7}$ | $\frac{5}{14}$ | $\frac{5}{14}$ | 0 |
| $P^3$  | 0.4  | 0.5  | 0.1  | 0    |

However, the above example illustrates that the mentioned generalised notions of conditioning do not satisfy the generalisation of the preservation property (1). If $\phi$ and $\psi$ are allowed to range over $SL_{PROB}$, then $\phi$ is no longer guaranteed to be accepted in $(P_\phi)_\psi$. Consider, for example, $P = P^1$ from example 3, $\phi : P(p \wedge q) = \frac{2}{7}$, and $\psi : P(p) = 0.9$. We conclude that conditioning does not allow a preservative change of the ignorant belief state into an arbitrary belief state.

In the context of partial probabilistic belief models for $L$, the ignorant belief state is represented by $PROB(SL)$. It is easy to see that any partial probabilistic belief model determined by a subset $S$ of $SL_{PROB}$ can be obtained from $PROB(SL)$ by constraining with the sentences of $S$. In other words, any probabilistic belief set $S$ for $L_{PROB}$ can be obtained by constraining the ignorant belief state $Cn_{PROB}(\emptyset)$. Moreover, constraining satisfies the appropriate generalisation of the preservation property (1):

**Proposition 5** *Let $S = \langle \Pi \rangle$ be a (probabilistic) belief set for $L_{PROB}$, and let $\psi, \phi \in SL_{PROB}$ such that $S, \phi,$ and $\psi$ are jointly consistent. Then*

$$\phi \text{ is accepted in } (\Pi_{\&\phi})_{\&\psi}.$$

Of course, one cannot constrain ignorance to partial probabilistic belief models which are not determined by a subset $S$ of $SL_{PROB}$. Since a similar situation arises in the case of expanding belief sets and possible worlds models of a language with infinitely many proposition letters, we do not consider this to be an essential difference between constraining and expansion.

The situation is much worse for conditioning, since very few partial probabilistic belief models can be obtained from $PROB(SL)$ by extended Bayesian conditioning. Given our previous deliberations, it may be natural to consider set-extensions of the discussed generalisations of Bayesian conditioning. For example, for any $\phi \in SL_{PROB}$, one can define $\Pi_\phi = \{P_\phi : P \in \Pi\}$, where $P_\phi$ is the minimum cross entropy update of $P$ with $\phi$. The following example shows that this operation does not satisfy the preservation property.

**Example 4** *Assume that $p$ and $q$ are the proposition letters of $L$, and let $\Pi = PROB(SL)$, $\phi : P(p \wedge q) = 0.5$, and $\psi : P(p) = 0.5$. Then $\Pi$, $\phi$, and $\psi$ are jointly*

*consistent, but it is not the case that $\phi$ is accepted in $(\Pi_\phi)_\psi$. The following table shows what happens to $P_{un}$ during the updates.*

|        | $p \wedge q$ | $p \wedge \neg q$ | $\neg p \wedge q$ | $\neg p \wedge \neg q$ |
|--------|------|------|------|------|
| $P_{un}$ | 0.25 | 0.25 | 0.25 | 0.25 |
| $(P_{un})_\phi$ | 0.5 | $\frac{1}{6}$ | $\frac{1}{6}$ | $\frac{1}{6}$ |
| $((P_{un})_\phi)_\psi$ | 0.375 | 0.125 | 0.25 | 0.25 |

*Since $P_{un} \in PROB(SL)$, we have $((P_{un})_\phi)_\psi \in (\Pi_\phi)_\psi$. Hence we cannot have $(\Pi_\phi)_\psi(p \wedge q) = 0.5$.*

Notice that the example shows that the preservation property is not even satisfied by the set-extension of binary Jeffrey conditioning. We conclude that (iterated) expansion has a certain property, namely the possibility of reaching every (definable) belief state from ignorance in a preservative manner, which is also possessed by constraining, but not by conditioning.

## 5   Conditioning versus Constraining

So far, we argued that conditioning or constraining are two different kinds of operations on belief states, where constraining is the proper probabilistic notion of expansion, since it is the expansion of probabilistic belief sets, and it can model preservative changes from the ignorant belief state to an arbitrary belief state. A decrease in ignorance is the main effect of constraining, whereas conditioning is primarily aimed at reducing uncertainty.

We have left open the question which of the two operations one should use when receiving information. We will discuss this matter using a well-known example deriving from (Smets, 1988).

**Example 5 (The three assassins)** *Mr. Jones has been murdered by one of the assassins Peter, Paul, and Mary under orders of Big Boss, who has chosen between these three possible killers as follows. He decided between a male and a female killer by means of tossing a fair coin. A male killer was chosen in case the coin landed heads. Otherwise, a female killer was chosen. No information is available on how he decided between the two male assassins in case the coin landed heads.*

*Based on the information above, it seems reasonable to say that the possibility of the killer being male and that of the killer being female are equally likely. Now suppose that you learn that at the time of the murder, Peter was at the police station, where he was questioned about some other crime. So you can rule out Peter as the killer. How should this new evidence be modelled? In particular, is it still equally likely for the killer to be male or female?*

To formalise this example, let $L$ be the language with the three propositional letters $p$, $q$, and $r$, where $p$ : Peter is the killer, $q$ : Paul is the killer, and $r$ : Mary is the killer. Given that exactly one of the assassins murdered Jones, only three atoms remain possible, namely, $\alpha = p \wedge \neg q \wedge \neg r$, $\beta = \neg p \wedge q \wedge \neg r$, and $\gamma = \neg p \wedge \neg q \wedge r$.

Adding the information that a fair coin toss decided the choice between a male and female killer leads to the partial probabilistic belief state $\Pi = \{P \in PROB(SL) : P(\alpha \vee \beta) = 0.5, P(\gamma) = 0.5\}$. This agrees with the interpretation in Dempster-Shafer theory, where the information is encoded in the mass function $m$ given by $m(\alpha \vee \beta) = 0.5, m(\gamma) = 0.5$, which induces a belief function $Bel$ such that $Bel$ is the lower envelope of $\Pi$. Strict Bayesians will opt for the probability function $P$ given by $P(\alpha) = 0.25$, $P(\beta) = 0.25$, and $P(\gamma) = 0.5$, which is the 'least informative' member of $\Pi$.

How to take account of the information $\neg p$ that Peter is not the killer? Strict Bayesians use Bayesian conditioning and arrive at $P_{\neg p}$ given by $P_{\neg p}(\beta) = \frac{1}{3}$, and $P_{\neg p}(\gamma) = \frac{2}{3}$, which implies that it is twice as likely for the killer to be female than to be male. However, as argued by (Smets, 1988) and (Halpern and Fagin, 1992), the 'least informative' prior $P$ on which this answer is based makes some (unjustified) assumptions about how the choice between Peter and Paul is made.

Starting from the partial probabilistic belief state $\Pi$, one can use both constraining and (extended) conditioning. Constraining $\Pi$ with $\neg p$ (or $P(\neg p) = 1$) gives $\Pi_{\&\neg p} = \{P \in PROB(SL) : P(\beta) = 0.5, P(\gamma) = 0.5\}$, which implies that the possibility of the killer being male and that of the killer being female are still equally likely. This answer completely agrees with the answer given be Dempster's rule of conditioning in Dempster-Shafer theory, and is defended by Smets (Smets88).

Conditioning $\Pi$ with $\neg p$ gives $\Pi_{\neg p} = \{P \in PROB(SL) : 0 \leq P(\beta) \leq 0.5, 0.5 \leq P(\gamma) \leq 1, P(\beta \vee \gamma) = 1\}$, which implies that the possibility of the killer being female is at least as likely as that of the killer being male. This answer, which is defended by (Halpern and Fagin, 1992), agrees with our intuition that finding out that Peter has an alibi makes it less likely that the coin landed heads to a degree which equals one's degree of belief that Peter would have been chosen in case of heads. The ignorance concerning $P(p|p \vee q)$ makes it impossible to justify a specific answer.

The above is related to a distinction discussed in (Dubois and Prade, 1997) between specific information, or factual evidence, which concerns a particular case at hand, and general information, or generic, background knowledge, which pertains to a class of sit-uations. Constraining is applicable in case of (general) information referring to the prior probabilities, and (specific) information about the case at hand should be incorporated using (extended) conditioning.

For example, learning Peter's alibi provides specific information, whereas a report of an undercover agent saying that Big Boss decided to choose Paul in the event that a male killer had to be chosen constitutes general information. In both cases one learns $\neg p$, but if it is specific information one has to condition on this event, whereas one has to use constraining in case the information is general.

The distinction between specific and general information is hard to make precise in general, but we find the following argument that Peter's alibi is specific, and not general, quite convincing.

**Example 6** *Assume that in the context of example 5, you received a report of an undercover agent saying that Big Boss, in the event he has to choose a male killer, decides between Peter and Paul by means of a (second) fair coin toss. Further assume that this report is completely reliable, but that you read it after you learned about Peter's alibi.*

The report of the undercover agent tells you that $P(p)/(P(p) + P(q)) = 0.5$. Simply adding this constraint to $\Pi_{\neg p}$ (as previously defined) results in $\{P \in PROB(SL) : P(\gamma) = 1\}$, which is counterintuitive, and adding the constraint to $\Pi_{\&\neg p}$ is not possible at all, since it leads to an inconsistency. However, the constraint from the report is a constraint on the *prior* probabilities, not on the probabilities obtained after incorporating the evidence of Peter's alibi.

Adding the constraint to $\Pi$ results in $\{P\}$, where $P$ is the given by $P(\alpha) = 0.25$, $P(\beta) = 0.25$, and $P(\gamma) = 0.5$. Conditioning on the evidence of Peter's alibi gives $\{P_{\neg p}\}$, which agrees with the answer given by strict Bayesians in the original example, since the probability function $P$ is chosen by the strict Bayesians even without the information from the undercover agent. Notice, however, that it is not possible to use constraining to incorporate Peter's alibi in the belief state $\{P\}$. This supports our opinion that Peter's alibi is specific information which calls for conditioning and not for constraining.

It can be shown that the order among conditionings or among constrainings does not matter. More precisely, if for some $P \in \Pi$, $P(\phi \wedge \psi) > 0$, then $(\Pi_\phi)_\psi = \Pi_{\phi \wedge \psi} = (\Pi_\psi)_\phi$, and if $\phi \wedge \psi$ is satisfied by some $P \in \Pi$, then $(\Pi_{\&\phi})_{\&\psi} = \Pi_{\&(\phi \wedge \psi)} = (\Pi_{\&\psi})_{\&\phi}$.

In contrast, $(\Pi_\phi)_{\&\psi} = (\Pi_{\&\psi})_\phi$ is *not valid*, as shown by the analysis of the example above. Constrain-ing should always be performed *before* conditioning, which implies that after conditioning one should not forget about the original belief state, since general constraints should be added to this original belief state.

Usually, one obtains additional information from observations of some aspects of the particular case at hand. This kind of information tends to be specific. But in some situations it might be reasonable to search for general information. For example, a chief of police, having the information described in the original example 5 of the three assassins, may conclude that he should assign at least as many detectives to investigate Mary as to investigate Paul. However, to determine an optimal division of the available detectives, he might try to reduce his ignorance by ordering an undercover agent to acquire information about the choice between Peter and Paul.

We conclude that the most common kind of probabilistic evidence is specific evidence about the case at hand, and should be incorporated by means of (extended) conditioning. Some evidence may be of a general nature and may reduce the ignorance concerning prior probabilities. Such evidence calls for constraining.

# 6  Conclusion

We argued that conditioning can best be viewed as a type of belief change different from expansion, revision, and contraction. This difference becomes most clear in the context of belief states represented by partially determined probability functions, which allow the representation of both ignorance and uncertainty. In this context, there are several operations agreeing with expansion on belief sets. Of these, constraining has more right to be called probabilistic expansion, than conditioning has, although the latter is often chosen in the literature.

The principal result of constraining is a decrease in ignorance, whereas conditioning is aimed at reducing uncertainty. General evidence reducing ignorance concerning the prior probabilities calls for constraining, but the most common probabilistic evidence concerns the particular case at hand, and should be modelled by means of conditioning. Constraining should always be applied *before* conditioning, since it represents evidence concerning the *prior* probabilities.

Conditioning does not make sense in the context of belief sets, which do not represent uncertainty, but only ignorance, just as expansion does not make sense in the context of belief states represented by probability functions, which do not represent ignorance, but only uncertainty. Our distinction between expansion and conditioning calls into question the treatments of probabilistic revision which start from the assumption that conditioning is the correct notion of probabilistic expansion.

## Acknowledgements

The investigations were carried out as part of the PIONIER-project Reasoning with Uncertainty, subsidized by the Netherlands Organization of Scientific Research (NWO), under grant pgs-22-262.

## References

D. Dubois and H. Prade, Belief change and possibility theory, in: P. Gärdenfors, ed., *Belief Revision* (Cambridge U.P., Cambridge, 1992) 142–182.

D. Dubois and H. Prade, Focusing vs belief revision: a fundamental distinction when dealing with generic knowledge, in D.M. Gabbay et al., eds., *Qualitative and Quantitative Practical Reasoning. Proceedings ECSQARU-FAPR'97*, LNCS 1244 (Springer, Berlin, 1997) 96–107.

R. Fagin, J.Y. Halpern and N. Megiddo, A logic for reasoning about probabilities, in *Proc. 3rd IEEE Symp. on Logic in Computer Science* (1988) 277–191.

P. Gärdenfors, *Knowledge in Flux. Modeling the Dynamics of Epistemic States* (MIT Press, Cambridge MA, 1988).

P. Gärdenfors, Belief revision: An introduction, in: P. Gärdenfors, ed., *Belief Revision* (Cambridge U.P., Cambridge, 1992) 1–28.

A.J. Grove and J.Y. Halpern, Updating sets of probabilities, *Proceedings 14th Conference on Uncertainty in AI* (1998) 173-182.

J.Y. Halpern and R. Fagin, Two views of belief: belief as generalized probability and belief as evidence, *Artificial Intelligence* 54 (1992) 275–317.

G.J. Klir, Measures of uncertainty in the Dempster-Shafer theory of evidence, in: R.R. Yager, J. Kacprzyk, and M. Fedrizzi, eds., *Advances in the Dempster-Shafer Theory of Evidence* (Wiley, New York, 1994) 35–49.

P. Smets, Belief functions versus probability functions, in: B. Bouchon, L. Aitt, R.R. Yager, eds., *Uncertainty and Intelligent Systems*, LNCS 313 (Springer, Berlin, 1988) 17–24.

F. Voorbraak, Probabilistic belief expansion and conditioning, *ILLC Research Report LP-96-07*, (ILLC, University of Amsterdam, 1996). (Slightly revised version submitted to Journal of Logic, Language and Information).